\documentclass[a4paper, 10pt, conference]{IEEEtran}
\IEEEoverridecommandlockouts
\usepackage{cite}
\usepackage{amsmath,amssymb,amsfonts}
\usepackage{graphicx}
\usepackage{textcomp}
\usepackage{adjustbox}
\usepackage{balance}
\usepackage{threeparttable}
\usepackage{soul}
\usepackage{verbatim}
\usepackage{url}
\usepackage{amssymb}
\usepackage{mathtools}
\usepackage{xparse}
\usepackage[percent]{overpic}
\usepackage{textcomp}
\usepackage{comment}
\usepackage{multirow}
\usepackage{array}
\usepackage{nicefrac} 
\usepackage{algorithm} 
\usepackage{algpseudocode} 
\usepackage[hidelinks]{hyperref}
\usepackage{siunitx, booktabs}
\hypersetup{
    colorlinks=true,
    linkcolor=black,
    filecolor=none,      
    urlcolor=blue,
    citecolor=black,
    pdftitle={Overleaf Example},
    pdfpagemode=FullScreen,
    }
\makeatletter
\newcommand{\raisemath}[1]{\mathpalette{\raiseMath{#1}}}%
\newcommand{\raiseMath}[3]{\raisebox{#1}[0pt][0pt]{$#2#3$}}

\makeatother

\NewDocumentCommand{\qbar}{O{0.5pt} O{-6.55pt}}{
	\ensuremath{\mathrlap{\raisemath{#2}{\hspace*{#1}{\mathchar'26\mkern-9mu}}} q}%
}

\NewDocumentCommand{\qbars}{O{0.5pt} O{-4.65pt}}{
	\ensuremath{\mathrlap{\raisemath{#2}{\hspace*{#1}{\mathchar'26\mkern-9mu}}} q}%
}

\NewDocumentCommand{\qbarc}{O{0.5pt} O{-5.2pt}}{
	\ensuremath{\mathrlap{\raisemath{#2}{\hspace*{#1}{\mathchar'26\mkern-9mu}}} q}%
}

\NewDocumentCommand{\pbar}{O{-1.5pt} O{-6.65pt}}{
	\ensuremath{\mathrlap{\raisemath{#2}{\hspace*{#1}{\mathchar'26\mkern-9mu}}} p}%
}

\newcommand{\bs}[1]{\boldsymbol{#1}}  
\newcommand{\ts}[1]{\text{#1}}

\renewcommand{\baselinestretch}{0.8573}  

\def\BibTeX{{\rm B\kern-.05em{\sc i\kern-.025em b}\kern-.08em
    T\kern-.1667em\lower.7ex\hbox{E}\kern-.125emX}}
\begin{document}

\title{\LARGE \bf A \mbox{Lyapunov-Based} Switching Scheme for Selecting the Stable \mbox{Closed-Loop} Fixed \mbox{Attitude-Error} Quaternion During Flight \\
\thanks{This work was supported by the Joint Center for Aerospace Technology Innovation (JCATI) through \mbox{Award\,172}, the Washington State University (WSU) Foundation and the Palouse Club through a Cougar Cage Award to \mbox{N.\,O.\,P\'erez-Arancibia}, and the WSU Voiland College of Engineering and Architecture through a \mbox{start-up} package to \mbox{N.\,O.\,P\'erez-Arancibia}.}
\thanks{Corresponding authors' email: {\tt francisco.goncalves@wsu.edu} (\mbox{F.\,M.\,F.\,R.\,G.}); {\tt n.perezarancibia@wsu.edu} (\mbox{N.\,O.\,P.-A.}).}%
}

\author{
\vspace{-2ex}
\mbox{Francisco M. F. R. Gon\c{c}alves}\textsuperscript{1}, Ryan M. Bena\textsuperscript{2}, Konstantin I. Matveev\textsuperscript{1}, and \mbox{N\'estor O. P\'erez-Arancibia}\textsuperscript{1}\\\\

\small \textsuperscript{1}School of Mechanical and Materials Engineering, Washington State University, Pullman, WA\,$99164$-$2920$, USA\\
\textsuperscript{2}Department of Mechanical and Civil Engineering, California Institute of Technology, Pasadena, CA\,$91125$-$2100$, USA
}

\maketitle
\thispagestyle{empty}
\pagestyle{empty}

\begin{abstract}
We present a switching scheme, which uses both the \textit{\mbox{attitude-error} quaternion} (AEQ) and the \mbox{angular-velocity} error, for controlling the rotational degrees of freedom of an \textit{uncrewed aerial vehicle} (UAV) during flight. In this approach, the proposed controller continually selects the stable \textit{\mbox{closed-loop}} (CL) equilibrium AEQ corresponding to the smallest cost between those computed with two \mbox{energy-based} Lyapunov functions. To analyze and enforce the stability of the CL switching dynamics, we use basic nonlinear theory. This research problem is relevant because the selection of the stable CL equilibrium AEQ directly determines the power and energy requirements of the controlled UAV during flight. To test and demonstrate the implementation, suitability, functionality, and performance of the proposed approach, we present experimental results obtained using a \mbox{$\bs{31}$-gram} quadrotor, which was controlled to execute \mbox{high-speed} yaw maneuvers in flight. These flight tests show that the proposed switching controller can respectively reduce the control effort and rotational power by as much as $\bs{49.75}$\,\% and $\bs{28.14}$\,\%, on average, compared to those corresponding to an \mbox{often-used} benchmark controller.
\end{abstract}

\section{Introduction}
\label{Section01}
\mbox{Quaternion-based} attitude representations of rigid objects in space are widely used to design and implement robust attitude controllers because they offer advantages compared to those based on Euler angles and rotation matrices, including numerical robustness and reduced computational complexity~\cite{Ying_ICRA_2016, Ying_IEEE_TCST_2020, Ying_ACC_2017, Ying_IROS_2018, Ying_ICRA_2019, Ying_Automatica_2024,Goncalves2024ICRA,Bena_RAL_Perception,BenaMPPC2022,Bena2023Yaw,calderon2019control,yang2019bee, Salcudean, Wie_Sign, Thienel_sign, Kristiansen,Fjellstad_sign,bhat2000topological,Schlanbusch2010Choosing,Mayhew_Robust,QuatAutomatica,comparisonEulerQuaternion}. For example, the use of quaternions eliminates the problems associated with the kinematic singularity known as \textit{gimbal lock} and, additionally, enables relatively faster computation due to the compact storage of attitude information~\cite{wen1991attitude, EulerQuaternionEfficiency}. However, the \mbox{so-called} \textit{quaternion ambiguity} issue induces representational and \mbox{decision-making} problems when quaternions are used to describe the dynamics of aerial robotic systems operating in a \mbox{\textit{closed-loop}} (CL) configuration. For instance, we show in~\mbox{\cite{calderon2019control,yang2019bee,BenaMPPC2022,Bena2023Yaw}} that the nonlinear dynamics of a micro \mbox{flapping-wing} \textit{uncrewed aerial vehicle} (UAV) can be robustly stabilized using a \mbox{Lyapunov-based} \mbox{continuous-time} \textit{linear \mbox{time-invariant}} (LTI) controller; but, interestingly, the resulting CL system has two different fixed points that represent exactly the same kinematic state, one stable and the other unstable. Furthermore, the stability properties of these two equilibria can be interchanged by simply changing the sign of a controller gain. This apparent paradox is explained by the fact that, after closing the loop, the controller becomes an integral part of the resulting CL dynamic system and, therefore, its interaction with the two mathematical representations of the same physical equilibrium generates two distinct CL dynamical behaviors. As discussed in Section\,\ref{Section03}, this phenomenon can greatly affect the performance of a controlled UAV during \mbox{high-speed} flight.

The ability to minimize both the tracking error and power consumption during flight are essential requirements for UAVs to efficiently perform \mbox{energy-demanding} tasks that necessitate time coordination, last long periods of time ($\sim$\,hours), and involve \mbox{high-speed} rotations, such as search and rescue missions, military operations, and wildlife monitoring~\cite{considerations}. This need for reliability and endurance motivated the research presented in this paper. One conceivable application of the method discussed here is \mbox{insect-flight} tracking for biological research. For instance, we can imagine a flying microrobot, such as the Bee\textsuperscript{++} presented in~\cite{Bena2023Yaw}, embedded in a flying bee colony when commanded to rapidly rotate during flight in order to change the specific insect that it is visually following at a given instant. To start this research, we hypothesized that, depending on both the controlled UAV's angular velocity and orientation when executing a rotational maneuver, in some instances it is beneficial to interchange the stability properties of the two equilibria of the CL system in real time, according to a \textit{performance figure of merit} (PFM). This hypothesis was inspired by a phenomenon known as \textit{unwinding behavior} and the methods employed to ensure its avoidance~\cite{bhat2000topological}. 

For \mbox{quaternion-based} attitude controllers of the type considered in this paper, the stability properties of the two CL fixed points are directly determined by the sign of a single gain in the feedback control law. For this purpose, we and other researchers have, in the past, simply used a signum function~\cite{Salcudean, Wie_Sign, Thienel_sign, Kristiansen, Fjellstad_sign, Bena_RAL_Perception, Bena2023Yaw, BenaMPPC2022, yang2019bee, calderon2019control}. For example, we included the sign of the scalar part of the \textit{\mbox{attitude-error} quaternion} (AEQ) in the control law presented in~\cite{BenaMPPC2022,Bena2023Yaw,calderon2019control,yang2019bee}, which is a way to choose as the stable CL equilibrium the fixed AEQ associated with the shortest rotational path required to make the instantaneous attitude control error equal to zero. Exactly the same approach was used in the definition of the backstepping flight control law presented and discussed in~\cite{Kristiansen}. Along the same lines, the work presented in~\cite{Schlanbusch2010Choosing} introduced a set of \mbox{\textit{a-priori}} rules to select as the stable CL equilibrium the fixed AEQ that heuristically minimizes a PFM that accounts for the control effort during flight; the research presented in~\cite{QuatAutomatica} developed a hybrid controller that takes into account both the \mbox{angular-velocity} error and AEQ in a CL \mbox{fixed-point} switching scheme; and, the research presented in~\cite{Mayhew_Robust} addressed the problem by introducing two hybrid control schemes derived from \mbox{energy-based} and backstepping Lyapunov functions, respectively. More recently, we introduced a new \mbox{model-predictive} method that uses a \mbox{first-principles-derived} mathematical description of the controlled UAV's attitude dynamics to predict and select the most \mbox{cost-efficient} stable CL equilibrium AEQ, according to a PFM, during flight~\cite{Goncalves2024ICRA}.

Except for the case we presented in~\cite{Goncalves2024ICRA}, none of the research efforts listed above validated the efficacy and robustness of the proposed methods through \mbox{real-time} experiments, and only numerical simulations were used for testing purposes. We believe that the thorough collection of experimental data is an essential step in the development of new algorithms for \mbox{real-time} control. In this paper, we introduce a \mbox{Lyapunov-based} switching controller that considers both the instantaneous AEQ and \mbox{angular-velocity} error when making decisions regarding the selection of the stable CL fixed AEQ. As part of the proposed approach, we present a method to enforce the stability of the CL system for a switching law specified in terms of the difference between two Lyapunov functions respectively associated with the two subsystems composing the CL switching dynamics. To test and demonstrate the functionality, suitability, and performance of the proposed approach, we present experimental results collected using a \mbox{$31$-gram} quadrotor during flight. Overall, the results presented here represent relevant theoretical and experimental contributions to the design, analysis, and implementation of controllers capable of selecting the stable CL equilibrium AEQ during \mbox{real-time} operation.

The rest of the paper is organized as follows. Section\,\ref{Section02} describes the kinematics and dynamics of the quadrotor flyer used to exemplify the process of controller design and in \mbox{real-time} flight experiments. Also, this section presents and explains a \mbox{Lyapunov-based} continuous controller used for analysis and as a starting point for research. Section\,\ref{Section03} discusses the \mbox{quaternion-based} representation of the controlled UAV's CL \mbox{attitude-error} dynamics corresponding to the continuous controller presented in Section\,\ref{Section02}, and explains
the performance problem that arises from the \mbox{quaternion-representation} ambiguity issue. Section\,\ref{Section04} presents the basic architecture, synthesis method, and stability analysis of the proposed switching control scheme. Section\,\ref{Section05} presents and discusses experimental results obtained through flight tests with \mbox{fast-changing} attitude references to provide evidence of the suitability and functionality of the proposed control approach. Last, Section\,\ref{Section06} states some conclusions and discusses possible directions for future research. 

\vspace{2ex}
\textit{\textbf{Notation:}}
\begin{enumerate}
\item\,$\mathbb{R}$ denotes the set of real numbers. Consistently, $\mathbb{R}^3$ denotes the set of real triplets.
\item\,$\mathcal{S}^3$ denotes the set of unit quaternions.
\item\,The variable $t$ denotes continuous time.
\item\,Italic lowercase symbols denote scalars, e.g., $p$; bold lowercase symbols denote vectors, e.g., $\bs{p}$; bold uppercase symbols denote matrices, e.g., $\bs{P}$; and bold crossed lowercase symbols denote quaternions, \mbox{e.g., $\bs{\pbar}$}.
\item\,We denote differentiation with respect to time using the dot operator, e.g., \mbox{$\dot{p} = \frac{dp}{dt}$}.
\item\,The symbols $\times$ and $\otimes$ denote the vector \mbox{cross-product} and quaternion product operations, respectively.
\item\,The operator $\left[\,\cdot\,\right]^T$ denotes the transposition of a matrix.
\item\,The operators $\lambda_{\ts{min}}\{\cdot\}$ and $\lambda_{\ts{max}}\{\cdot\}$ extract the minimum and maximum eigenvalues of a square matrix.
\item\,The operator \mbox{$\| \cdot \|_2$} computes the $2$-norm of a vector.
\item\,The operator $\ts{sgn} \left\{\,\cdot\,\right\}$ extracts the sign of a real scalar.
\end{enumerate}
\begin{figure}[t!]
\vspace{1ex}
\begin{center}
\includegraphics[width=0.48\textwidth]{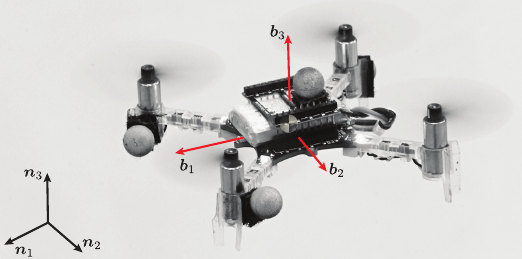}
\end{center}
\vspace{-2ex}     
\caption{\textbf{Photograph of the UAV used in the flight tests, the Crazyflie\,$\boldsymbol{2.1}$, with reflective markers attached to its body.} Here, \mbox{$\bs{\mathcal{B}} = \left\{\bs{b}_1,\bs{b}_2,\bs{b}_3 \right\}$}, with its origin coinciding with the UAV's center of mass, is the \mbox{body-fixed} frame of reference; \mbox{$\bs{\mathcal{N}} = \left\{\bs{n}_1,\bs{n}_2,\bs{n}_3 \right\}$} is the inertial frame of reference. \label{FIG01}} 
\vspace{-2ex}
\end{figure}

\section{Dynamical Model of the Controlled UAV and Continuous Controller Used for Analysis}
\label{Section02}
\subsection{Reduced-Complexity Rigid-Body Dynamics of the UAV}
\label{Subsection02A}
We first present a \mbox{reduced-complexity} dynamical model of the controlled UAV that is simple but sufficiently accurate for synthesizing and implementing feedback controllers. Accordingly, \mbox{Fig.\,\ref{FIG01}} shows the \mbox{body-fixed} and inertial frames, \mbox{$\bs{\mathcal{B}} = \left\{\bs{b}_1,\bs{b}_2,\bs{b}_3 \right\}$} and \mbox{$\bs{\mathcal{N}} = \left\{\bs{n}_1,\bs{n}_2,\bs{n}_3 \right\}$}, used to describe the kinematics and dynamics of the system. Here, the origin of \mbox{$\bs{\mathcal{B}}$} is defined to coincide with the \textit{center of mass} (COM) of the UAV. As discussed in~\cite{BenaMPPC2022}~and~\cite{Bena2023Yaw}, directly from Newton's and d'Alembert's laws, it follows that 
\begin{align}
\begin{split}
\bs{\dot{r}} &= \bs{v},  \\
\bs{\dot{v}} &= \frac{f_{\ts{a}}}{m}\bs{b}_3 - g\bs{n}_3, \\
\bs{\dot{\qbar}} &= \frac{1}{2} \bs{\qbar}\otimes
\begin{bmatrix}
0 \\
\bs{\omega} 
\end{bmatrix},\\
\bs{\dot{\omega}} &=  \bs{J}^{-1}\left(\bs{\tau}-\bs{\omega}\times \bs{J}\bs{\omega}\right),
\label{EQ01}
\end{split}
\end{align}
where $\bs{r}$ is the instantaneous position of the origin of \mbox{$\bs{\mathcal{B}}$} relative to the origin of $\bs{\mathcal{N}}$; $\bs{v}$ is the velocity of the UAV's COM relative to the origin of $\bs{\mathcal{N}}$; $f_{\ts{a}}$ is the magnitude of the total \mbox{non-gravitational} (\mbox{actuation-induced}) force acting on the UAV, which is assumed to be perfectly aligned with $\bs{b}_3$ because of the configuration of the UAV; $m$ is the total mass of the UAV; $g$ is the acceleration of gravity; $\bs{\qbar}$ is a unit quaternion that stores the attitude information of the UAV during flight; $\bs{\omega}$ is the angular velocity of $\bs{\mathcal{B}}$ relative to $\bs{\mathcal{N}}$, written with respect to $\bs{\mathcal{B}}$; $\bs{J}$ is the inertia matrix of the UAV's rigid body, which is written with respect to $\bs{\mathcal{B}}$ and, therefore, constant; and, $\bs{\tau}$ is the total \mbox{actuation-induced} torque acting on the UAV during flight. As explained in~\cite{yang2019bee,BenaMPPC2022,Bena2023Yaw,calderon2019control}, $f_{\ts{a}}$ and $\bs{\tau}$ are the inputs of the system, generated in closed loop by the propellers of the flyer as a result of the commands computed by an onboard feedback controller. As it is well known~\cite{KuipersQuaternions,markley2014fundamentals,AltmannRotQuat}, the attitude unit quaternion has the form
\begin{align}
\bs{\qbar} = 
\left[
\begin{array}{c}
~\cos \frac{\rm{\Phi}}{2} \\
\vspace{-2ex}
\\
\bs{u} \sin \frac{\rm{\Phi}}{2}
\end{array}
\right]\hspace{-0.3ex},
\label{EQ02}
\end{align}  
which, in accordance with Euler's rotation theorem~\cite{KuipersQuaternions}, represents the orientation of $\bs{\mathcal{B}}$ with respect to $\bs{\mathcal{N}}$ as a single rotation of magnitude $\Phi$ about the unit axis $\bs{u}$. 

\subsection{Continuous Attitude Feedback Controller Used for Analysis}
\label{Subsection02B}
In this section, we provide an overview of the \mbox{model-based} continuous attitude feedback controller used for analysis and as a starting point of the research presented in this paper. This controller is based on ideas introduced to control spacecrafts and was initially developed to fly \mbox{bee-inspired} flapping-wing microrobots~\cite{yang2019bee,calderon2019control,BenaMPPC2022,Bena2023Yaw}. When controlling the attitude of a flyer, a main objective is minimizing the attitude error during flight. Using quaternions, this control error is defined as
\begin{align}
\bs{\qbar}_\text{e} = \bs{\qbar}^{-1} \otimes \bs{\qbar}_\text{d},
\label{EQ03}
\end{align}
where $\bs{\qbar}_\ts{d}$ stores the desired instantaneous attitude of $\bs{\mathcal{B}}$ and $\bs{\qbar}$ stores the measured instantaneous attitude of $\bs{\mathcal{B}}$. As explained in~\cite{BenaMPPC2022}, the AEQ \mbox{$\bs{\qbar}_{\ts{e}} = \left[ m_{\ts{e}} \,\, \bs{n}_{\ts{e}}^T \right]^T$} represents the attitude of the \mbox{\textit{desired}} \mbox{body-fixed} frame, $\bs{\mathcal{B}}_\ts{d}$, relative to $\bs{\mathcal{B}}$, with \mbox{$m_{\ts{e}} = \cos{\frac{\Phi_{\ts{e}}}{2}}$} and \mbox{$\bs{n}_{\ts{e}} = \bs{u}_{\ts{e}}\sin{\frac{\Phi_{\ts{e}}}{2}}$}, where $\Phi_{\ts{e}}$ is the amount that $\bs{\mathcal{B}}$ must be rotated about the Euler axis $\bs{u}_{\ts{e}}$ to be aligned with $\bs{\mathcal{B}}_\ts{d}$. Accordingly, the desired \mbox{angular~velocity} of $\bs{\mathcal{B}}$ relative to $\bs{\mathcal{N}}$, written using $\bs{\mathcal{B}}_\ts{d}$ coordinates, is given by
\begin{align}
\left[
\begin{array}{c}
0 \\
\vspace{-2ex}
\\
\bs{\tilde{\omega}}_\text{d}
\end{array}
\right]
 = 2\bs{\qbar}_\ts{d}^{-1} \otimes \dot{\bs{\qbar}}_\text{d}.
\label{EQ04}
\end{align}
Note that the desired angular velocity of $\bs{\mathcal{B}}$ relative to $\bs{\mathcal{N}}$, written using $\bs{\mathcal{B}}$ coordinates, can be readily found by computing
\begin{align}
\bs{\omega}_\text{d} = \bs{R}^T \bs{R}_\ts{d} \tilde{\bs{\omega}}_\text{d},
\label{EQ05}
\end{align}
where $\bs{R}$ is the rotation matrix that transforms vectors from $\bs{\mathcal{B}}$ coordinates to $\bs{\mathcal{N}}$ coordinates and $\bs{R}_\ts{d}$ is the rotation matrix that transforms vectors from $\bs{\mathcal{B}}_{\ts{d}}$ coordinates to $\bs{\mathcal{N}}$ coordinates.

In the control approach considered here, the torque inputted to the attitude dynamical system, as specified in (\ref{EQ01}), is generated according to a law that explicitly depends on the AEQ, $\bs{\qbar}_\text{e}$; measured angular velocity of $\bs{\mathcal{B}}$, $\bs{\omega}$; desired angular velocity of $\bs{\mathcal{B}}$, $\bs{\omega}_{\ts{d}}$; and, its derivative, $\bs{\dot{\omega}}_{\ts{d}}$. Namely,
\begin{align}
\bs{\tau} = \bs{K}_{\bs{\qbars}} \bs{n}_{\ts{e}} + \bs{K}_{\bs{\omega}} \bs{\omega}_{\ts{e}} + \bs{J}\bs{\dot{\omega}}_\ts{d} + \bs{\omega}\times\bs{J}\bs{\omega},
\label{EQ06}
\end{align}
where $\bs{K}_{\bs{\qbars}}$ and $\bs{K}_{\bs{\omega}}$ are \mbox{controller-gain} \mbox{positive-definite} matrices; and, \mbox{$\bs{\omega}_{\ts{e}}= \bs{\omega}_{\ts{d}} - \bs{\omega}$} is the \mbox{angular-velocity} tracking error. In this law, the first two terms correspond to a \mbox{\textit{proportional--derivative}} (PD) structure, where the term \mbox{$\bs{K}_{\bs{\qbars}}\bs{n}_{\ts{e}}$} generates a torque control action aligned with the direction of $\bs{\qbar}_{\ts{e}}$'s Euler axis of rotation, $\bs{u}_{\ts{e}}$, with the objective of directly compelling the alignment of $\bs{\mathcal{B}}$ with $\bs{\mathcal{B}}_{\ts{d}}$. The term $\bs{K}_{\bs{\omega}} \bs{\omega}_{\ts{e}}$ is proportional to the \mbox{angular-velocity} tracking error, thus corresponding to a derivative action. The last two terms correspond to a feedforward structure. Specifically, assuming close tracking, the objective of $\bs{J}\bs{\dot{\omega}}_\ts{d}$ is to cancel $\bs{J}\bs{\dot{\omega}}$ in (\ref{EQ01}); the objective of \mbox{$\bs{\omega}\times\bs{J}\bs{\omega}$} is to linearize the fourth row in (\ref{EQ01}).

\section{CL Stability and Benchmark Controller}
\label{Section03}
\subsection{Fixed Points of the CL Attitude Dynamics of the System}
\label{Subsection03A}
By differentiating (\ref{EQ03}) and substituting the \textit{\mbox{right-hand} side} (RHS) of (\ref{EQ06}) into the last row of (\ref{EQ01}), we obtain the CL attitude dynamics of the system. Namely,
\begin{align}
\begin{split}
\bs{\dot{\qbar}}_{\ts{e}} &= \frac{1}{2}
\left[
\begin{array}{c}
    0 \\
    \bs{\omega}_{\ts{e}}
\end{array}
\right]
\otimes \bs{\qbar}_{\text{e}},\\
    \bs{\dot{\omega}}_{\ts{e}} 
    &= -\bs{J}^{-1} \left( \bs{K}_{\bs{\qbars}} \bs{n}_{\ts{e}} + \bs{K}_{\bs{\omega}} \bs{\omega}_{\ts{e}} \right). \\
\end{split}
\label{EQ07}
\end{align}
As a design choice and for analysis purposes, we select the system references, $\bs{\qbar}_\ts{d}$ and $\bs{\omega}_{\ts{d}}$, to be smooth and bounded functions of time. Also, as discussed in~\cite{BenaMPPC2022}, the RHS vector functions of this \mbox{seven-variable} \mbox{state-space} system are \mbox{Lipschitz} continuous almost everywhere. It is straightforward to show that (\ref{EQ07}) has two fixed points, one stable and another unstable. To find these two equilibria, we simply set \mbox{$\bs{\dot{\qbar}}_{\ts{e}} = \left[ 0\,\, 0\,\, 0\,\, 0 \right]^T$} and $\bs{\dot{\omega}}_{\ts{e}} = \left[0\,\, 0\,\, 0 \right]^T$, and then solve for $ \bs{\qbar}_{\ts{e}}$ and $\bs{\omega}_{\ts{e}}$, following the same logic employed in~\cite{BenaMPPC2022}. The first equilibrium state corresponds to \mbox{$\bs{\qbar}_{\ts{e}}^{\star} = \left[+1\,\, 0\,\, 0\,\, 0 \right]^T$} and \mbox{$\bs{\omega}_{\ts{e}}^{\star} =  \left[ 0\,\, 0\,\, 0 \right]^T$}; the second equilibrium state corresponds to \mbox{$\bs{\qbar}_{\ts{e}}^{\dagger} = \left[-1\,\, 0\,\, 0\,\, 0 \right]^T$} and \mbox{$\bs{\omega}_{\ts{e}}^{\star} =  \left[ 0\,\, 0\,\, 0 \right]^T$}. Note that these two fixed points represent the same \mbox{attitude-tracking} and \mbox{angular-velocity} errors; therefore, convergence to either of these two states represents convergence to the same physical state.

\subsection{Stability Analysis and Performance Problem}
\label{Subsection03B}
As shown in~\cite{BenaMPPC2022}, the stability of the first equilibrium point, \mbox{$\{\bs{\qbar}^{\star}_\ts{e}, \bs{\omega}^{\star}_\ts{e}\}$}, can be enforced by simply selecting the matrices $\bs{K}_{\bs{\qbars}}$ and $\bs{K}_{\bs{\omega}}$ to be positive definite. This result is reiterated formally in the following proposition.
\vspace{1ex}

\hspace{-2.2ex}\textbf{Proposition\,1.} \textit{Let the attitude and \mbox{angular-velocity} references}, $\bs{\qbar}_\ts{d}$ \textit{and} $\bs{\omega}_\ts{d}$, \textit{be smooth functions of time, and let} $\bs{K}_{\bs{\qbars}}$ \textit{and} $\bs{K}_{\bs{\omega}}$ \textit{be constant \mbox{positive-definite matrices}}. \textit{Then, the fixed point} $\{\bs{\qbar}^{\star}_\ts{e}, \bs{\omega}^{\star}_\ts{e}\}$ \textit{of the CL attitude dynamics specified by (\ref{EQ07}), with} \mbox{$\bs{\qbar}_{\ts{e}}^{\star} = \left[+1\,\, 0\,\, 0\,\, 0 \right]^T$} \textit{and} \mbox{$\bs{\omega}_{\ts{e}}^{\star} =  \left[ 0\,\, 0\,\, 0 \right]^T\hspace{-0.4ex}$}, \textit{is asymptotically stable.}

\vspace{1ex}
\hspace{-2.2ex}\textit{Proof.} See Section\,3.6 in~\cite{BenaMPPC2022}.
\vspace{1ex}

\hspace{-2.2ex}Similarly, the instability of the second equilibrium point, \mbox{$\{\bs{\qbar}^{\dagger}_\ts{e}, \bs{\omega}^{\star}_\ts{e}\}$}, can be readily shown using \textit{Lyapunov's indirect method} as stated in~\mbox{Theorem\,4.7}~of~\cite{khalil2002nonlinear}, when both matrices $\bs{K}_{\bs{\qbars}}$ and $\bs{K}_{\bs{\omega}}$ are chosen to be positive definite. This result is reiterated formally in the following proposition.
\vspace{1ex}

\hspace{-2.2ex}\textbf{Proposition\,2.} \textit{Let the attitude and \mbox{angular-velocity} references}, $\bs{\qbar}_\ts{d}$ \textit{and} $\bs{\omega}_\ts{d}$, \textit{be smooth functions of time and let matrices} $\bs{K}_{\bs{\qbars}}$ \textit{and} $\bs{K}_{\bs{\omega}}$ \textit{be constant and positive definite}. \textit{Then, the fixed point} $\{\bs{\qbar}^{\dagger}_\ts{e}, \bs{\omega}^{\star}_\ts{e}\}$ \textit{of the CL attitude dynamics specified by (\ref{EQ07}), with} \mbox{$\bs{\qbar}_{\ts{e}}^{\dagger} = \left[ -1\,\, 0\,\, 0\,\, 0 \right]^T$} \textit{and} \mbox{$\bs{\omega}_{\ts{e}}^{\star} =  \left[ 0\,\, 0\,\, 0 \right]^T\hspace{-0.4ex}$}, \textit{is unstable.}
\vspace{1ex}

\hspace{-2.2ex}\textit{Proof.}
See Appendix\,C in~\cite{BenaMPPC2022}.
\vspace{1ex}

As mentioned above and thoroughly discussed in~\cite{BenaMPPC2022} and \cite{Bena2023Yaw}, even though both equilibrium points of the CL system specified by (\ref{EQ07}), \mbox{$\{\bs{\qbar}^{\star}_\ts{e}, \bs{\omega}^{\star}_\ts{e}\}$} and \mbox{$\{\bs{\qbar}^{\dagger}_\ts{e}, \bs{\omega}^{\star}_\ts{e}\}$}, have opposite stability properties, they represent exactly the same physical state of the UAV during flight~\cite{bhat2000topological,Schlanbusch2010Choosing,Mayhew_Robust,QuatAutomatica}. This phenomenon is the direct result of the \mbox{quaternion-sign} ambiguity and, as discussed in~\cite{Goncalves2024ICRA}, can induce undesired \mbox{$2\pi$-rad} rotations during flight. A heuristic solution to this problem is to multiply the first term in the control law specified by (\ref{EQ06}) by the sign of $m_\ts{e}$. Thus, the torque inputted to close the loop becomes
\begin{align}
\bs{\tau}_{\ts{b}} = \ts{sgn}\{m_\ts{e}\}\bs{K}_{\bs{\qbars}} \bs{n}_{\ts{e}} + \bs{K}_{\bs{\omega}} \bs{\omega}_{\ts{e}} + \bs{J}\bs{\dot{\omega}}_\ts{d} + \bs{\omega}\times\bs{J}\bs{\omega}.
\label{EQ08}
\end{align}
In~\cite{Goncalves2024ICRA}, we thoroughly explain the effects of this modification. Even though there are numerous examples that provide evidence of the empirical effectiveness of this method, this approach does not necessarily produce the best performance in terms of the magnitude of the control signal or energy. Along these lines, we hypothesize that replacing the $\ts{sgn}\{m_\ts{e}\}$ term with a function that explicitly depends on both $\bs{\qbar}_\ts{e}$ and $\bs{\omega}_\ts{e}$ might be beneficial from a performance perspective in cases in which the component of $\bs{\omega}_\ts{e}$ aligned with $\bs{u}_{\ts{e}}$ is significant. For example, we can imagine the controlled UAV rotating at a high speed along a given direction, when the attitude reference, $\bs{\qbar}_\ts{d}$, abruptly and rapidly changes and the new shorter rotational path from $\bs{\mathcal{B}}$ to $\bs{\mathcal{B}}_{\ts{d}}$ is in the opposite direction to that before the attitude reference is changed. In this example, at high rotational speeds and considering inertia effects, it might not be the best decision from a performance perspective to abruptly change the direction of rotation. To address this issue, in Section\,\ref{Section04}, we propose a switching control law whose definition is based on two subsystems with the same equilibrium states but opposite stability properties. This new controller can be seen as an alternative to that specified by (\ref{EQ08}), which from this point onward, we refer to as the benchmark controller.

\section{A New Switching Attitude Controller}
\label{Section04}
\subsection{Definition of the Torque Control Signal}
\label{Section04A}
\begin{figure}[t!]
\vspace{1ex}
\begin{center}
\includegraphics[width=0.48\textwidth]{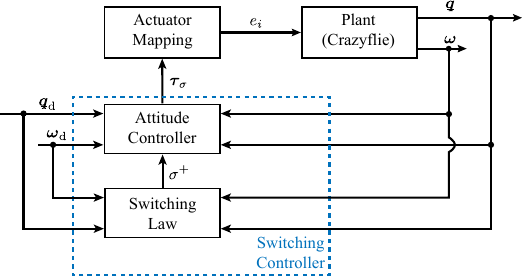}
\end{center}
\vspace{-2ex}
\caption{\textbf{Block diagram of the proposed switching controller.} The switching law receives the desired and measured attitude quaternions, $\bs{\qbarc}_\ts{d}$ and $\bs{\qbarc}$, and the desired and measured angular velocities, $\bs{\omega}_\ts{d}$ and $\bs{\omega}$. Then, it computes the switching signal \mbox{$\sigma^+ \in \left\{+1,-1\right\}$} and sends it to the attitude controller, which computes the control torque, $\bs{\tau}_{\hspace{-0.2ex}\sigma}$. Last, as explained in~\cite{Ying_ICRA_2016}, the actuator mapping receives $\bs{\tau}_{\hspace{-0.2ex}\sigma}$ and generates the \textit{pulse-width modulation} (PWM) voltage signals, $e_i$, \mbox{$i \in \left\{1,2,3,4 \right\}$}, that excite the four DC motors of the controlled UAV. \label{FIG02}}
\vspace{-3ex}
\end{figure}

We address the problem discussed in~Section\,\ref{Subsection03B} by introducing the control scheme shown in~Fig.\,\ref{FIG02}. Here, the torque inputted to the system specified by (\ref{EQ01}) is generated using a switching control law. This law is defined in terms of two subsystems as 
\begin{align}
\begin{split}
\bs{\tau}_{\hspace{-0.2ex}\sigma} = \sigma \bs{K}_{\bs{\qbars}}\bs{n}_{\ts{e},\sigma}  + \bs{K}_{\bs{\omega}} \bs{\nu}_{\sigma} + \bs{J}\left(\bs{\dot{\omega}}_{\ts{d}} + \sigma k_{\bs{n}} \bs{\dot{n}}_{\ts{e},\sigma} \right) + \bs{\omega}\times\bs{J}\bs{\omega},
\end{split}
\label{EQ09}
\end{align}
in which the variable \mbox{$\sigma \in \left\{+1,-1\right\}$}; \mbox{$\bs{n}_{\ts{e},\sigma}$} is the vector component of the instantaneous AEQ, \mbox{$\bs{\qbar}_{\text{e},\sigma}$}; \mbox{$\bs{\nu}_{\sigma}= \bs{\omega}_{\ts{e}} + \sigma k_{\bs{n}}\bs{n}_{\ts{e},\sigma}$}; and, \mbox{$0 < k_{\bs{n}} \in \mathbb{R}$}. The other variables were already defined in Section\,\ref{Subsection02B}. As seen, the main difference between the law defined by (\ref{EQ09}) and those specified by (\ref{EQ06}) and (\ref{EQ08}) is the inclusion of a term depending on $\bs{\dot{n}}_{\ts{e},\sigma}$, which, as explained in Section\,\ref{SubSection04C}, allows us to define a stabilizing time function $\Lambda$ for the switching condition based on both the AEQ and the \mbox{angular-velocity} error. We specify this switching condition using $\Lambda$ and a positive real constant $\delta$, according to
\begin{align}
\sigma^{+} =
\left\{
\begin{array}{ccc}
 ~~\sigma, & \textrm{if} & -\delta < \Lambda < \delta \\
 +1, &  \textrm{if} & ~~~~~~~\,\Lambda \geq \delta  \\
 -1, &  \textrm{if} & ~~~~~~~~~\,\Lambda \leq -\delta
\end{array}
\right..
\label{EQ10}
\end{align}
In this scheme, the condition specified by (\ref{EQ10}) is continuously checked; $\sigma$ denotes the current value of the sign parameter, initialized as \mbox{$\sigma=+1$}, and \mbox{$\sigma^{+}$} denotes the value of $\sigma$ to be used in~(\ref{EQ09}) at the next instant. We discuss the logic for the selection of $\Lambda$ and its form in \mbox{Section\,\ref{SubSection04C}}.

\subsection{Equilibrium Points of the New CL System} \label{Subsection04B}
To find the \mbox{state-space} representation of the CL system resulting from using the control law specified by (\ref{EQ09}), we simply plug (\ref{EQ09}) into the last equation of (\ref{EQ01}). This procedure yields
\begin{align}
\begin{split}
\bs{\dot{\qbar}}_{\ts{e},\sigma} &= \frac{1}{2}
\left[
\begin{array}{c}
    0 \\
    \bs{\nu}_{\sigma} - \sigma k_{\bs{n}}\bs{n}_{\ts{e},\sigma}
\end{array}
\right]
\otimes \bs{\qbar}_{\text{e},\sigma},\\
    \bs{\dot{\nu}}_{\sigma} 
    &= -\bs{J}^{-1} \left( \sigma \bs{K}_{\bs{\qbars}} \bs{n}_{\ts{e},\sigma} + \bs{K}_{\bs{\omega}} \bs{\nu}_{\sigma} \right), \\
\end{split}
\label{EQ11}
\end{align}
for \mbox{$\sigma \in \left\{+1,-1\right\}$}. According to this mathematical description, the state for each subsystem is $\left\{\bs{\nu}_{\sigma}, \bs{\qbar}_{\ts{e},\sigma} \right\}$. Note, however, that the state for the switching system as a whole remains to be the measured pair $\left\{\bs{\omega}_{\ts{e}}, \bs{\qbar}_{\ts{e}} \right\}$. To ensure asymptotic stability, we must force the energy of the CL system to decrease through each transition between the two control structures specified by (\ref{EQ09}). We formalize this idea in three steps. First, we find the fixed points of the subsystems that compose the switching dynamics specified by (\ref{EQ11}); then, we determine the conditions for their asymptotic stability; and, last, we use the logic in~\mbox{Theorem\,3.1} of~\cite{liberzon2003switching} and its proof to establish the asymptotic stability of the switching dynamics in~(\ref{EQ11}). As in the case discussed in Section\,\ref{Subsection03A}, we select the system references, $\bs{\qbar}_{\ts{d}}$ and $\bs{\omega}_{\ts{d}}$, to be smooth and bounded functions of time. Also, it can be shown that the vector functions on the right side of~(\ref{EQ11}) are Lipschitz continuous almost everywhere and that each \mbox{seven-variable} \mbox{state-space} subsystem has two equilibria; one stable and another unstable. To find the fixed points of each subsystem, we set \mbox{$\bs{\dot{\nu}}_{\sigma} = \left[0\,\, 0\,\, 0 \right]^T$} and \mbox{$\bs{\dot{\qbar}}_{\ts{e},\sigma} = \left[ 0\,\, 0\,\, 0\,\, 0 \right]^T$} and, then, use the set of algebraic equations
\begin{align}
\begin{split}
    -\frac{1}{2}\bs{n}_{\ts{e},\sigma}^T\bs{\nu}_{\sigma} +\frac{1}{2}\sigma k_{\bs{n}}\bs{n}_{\ts{e},\sigma}^T\bs{n}_{\ts{e},\sigma} &= 0,  \\
   - \frac{1}{2}\left[ \bs{n}_{\ts{e},\sigma} \times \bs{\nu}_{\sigma}  - m_{\ts{e},\sigma}\left(\bs{\nu}_{\sigma} - \sigma k_{\bs{n}}\bs{n}_{\ts{e},\sigma}\right)   \right] &= \bs{0}_{3\times1},\\
   -\bs{J}^{-1} \left( \sigma \bs{K}_{\bs{\qbars}} \bs{n}_{\ts{e},\sigma}+\bs{K}_{\bs{\omega}} \bs{\nu}_{\sigma} \right) &= \bs{0}_{3\times1},
\end{split}
\label{EQ12}
\end{align}
to solve for $\bs{\qbar}_{\ts{e},\sigma}$ and $\bs{\nu}_{\sigma}$. Specifically, through the same logic employed in~\cite{BenaMPPC2022}, it can be determined that each subsystem specified by (\ref{EQ11}) has two equilibrium points. The first point corresponds to the pair \mbox{$\bs{\nu}_{\sigma}^{\star} =  \left[ 0\,\, 0\,\, 0 \right]^T$} and \mbox{$\bs{\qbar}_{\ts{e},\sigma}^{\star} = \left[ +1\,\, 0\,\, 0\,\, 0 \right]^T$}, with \mbox{$\sigma \in \left\{+1, -1\right\}$}; the second point corresponds to \mbox{$\bs{\nu}_{\sigma}^{\star}$} and \mbox{$\bs{\qbar}_{\ts{e},\sigma}^{\dagger} = [-1\,\, 0\,\, 0\,\, 0]^T$}, with \mbox{$\sigma \in \left\{+1, -1\right\}$}.

\subsection{Stability of the Switching Dynamics}
\label{SubSection04C}
Now, we analyze and provide a method for ensuring the stability of the CL switching system using \mbox{Theorem\,3.1} in~\cite{liberzon2003switching} and \textit{Lyapunov's direct method} for nonautonomous systems as stated in~\mbox{Theorem\,4.9} of \cite{khalil2002nonlinear}.
\vspace{1ex}

\noindent\textbf{Proposition\,3.} \textit{Let the attitude and \mbox{angular-velocity} references}, $\bs{\qbar}_\ts{d}$ \textit{and} $\bs{\omega}_\ts{d}$, \textit{be smooth and bounded functions of time,} \textit{let} $\bs{K}_{\bs{\qbars}}$ \textit{and} $\bs{K}_{\bs{\omega}}$ \textit{be constant \mbox{positive-definite matrices},} \textit{and} \textit{let} $k_{\bs{n}}$ \textit{be a positive scalar.} \textit{Then, the fixed set} \mbox{$\{\sigma \bs{\qbar}^{\star}_\ts{e}, \bs{\nu}^{\star}_{\sigma}\}$}, \textit{for} \mbox{$\sigma \in \left\{+1,-1\right\}$}, \textit{of the CL \mbox{state-space} switching attitude dynamics specified by (\ref{EQ11}),} \textit{with} \mbox{$\bs{\qbar}_{\ts{e}}^{\star} = \left[ +1\,\, 0\,\, 0\,\, 0 \right]^T\hspace{-0.4ex}$} \textit{and} \mbox{$\bs{\nu}_{\sigma}^{\star} = \left[ 0\,\, 0\,\, 0 \right]^T\hspace{-0.4ex}$}, \textit{is asymptotically stable.}
\vspace{1ex}

\noindent\textit{Proof.}~First, for the subsystem corresponding to \mbox{$\sigma = +1$}, we show that the fixed point \mbox{$\{\bs{\qbar}^{\star}_{\ts{e},+1}, \bs{\nu}^{\star}_{+1} \}$} is asymptotically stable. In this case, we choose the \textit{Lyapunov function} (LF) defined as
\begin{align}
V_{+1}(\bs{\qbar}_{\ts{e}}, \bs{\nu}_{+1}) = \frac{1}{2} \bs{\nu}_{+1}^T\bs{K}_{\bs{\qbars}}^{-1}\bs{J}\bs{\nu}_{+1} + 2(1 - m_\ts{e}),
\label{EQ13}
\end{align}
where, by definition, \mbox{$\bs{\qbar}_{\ts{e}}(t) = \bs{\qbar}_{\ts{e},+1}(t)$} and, therefore, \mbox{$\bs{\qbar}^{\star}_{\ts{e}} = \bs{\qbar}^{\star}_{\ts{e},+1} = [+1\,\, 0\,\, 0\,\, 0]^T$}. It can be shown that \mbox{$V_{+1}(\bs{\qbar}_\ts{e},\bs{\nu}_{+1})$} is continuously differentiable, because both $\bs{\qbar}_\ts{d}$ and $\bs{\omega}_\ts{d}$ are smooth functions of time. Furthermore, by plugging \mbox{$\{\bs{\qbar}^{\star}_\ts{e},\bs{\nu}^{\star}_{+1}\}$} into (\ref{EQ13}), it follows that 
\begin{align}
V_{+1}(\bs{\qbar}_{\ts{e}}^{\star},\bs{\nu}_{+1
}^{\star})=0.
\label{EQ14}
\end{align}
Moreover, \mbox{$V_{+1}(\bs{\qbar}_\ts{e},\bs{\nu}_{+1})$} can be bounded according to
\begin{align}
    W_1(\bs{\qbar}_\ts{e},\bs{\nu}_{+1})\leq V_{+1}(\bs{\qbar}_\ts{e},\bs{\nu}_{+1}) \leq W_2(\bs{\qbar}_\ts{e},\bs{\nu}_{+1}),
    \label{EQ15}
\end{align}
with
\begin{align}
    W_1(\bs{\qbar}_\ts{e},\bs{\nu}_{+1})= \lambda_{\ts{min}}\left\{\bs{K}_{\bs{\qbars}}^{-1}\bs{J}\right\}\bs{\nu}_{+1}^T\bs{\nu}_{+1} + 2(1-m_{\ts{e}})
    \label{EQ16}
\end{align}
and
\begin{align}
    W_2(\bs{\qbar}_\ts{e},\bs{\nu}_{+1})= \lambda_{\ts{max}}\left\{\bs{K}_{\bs{\qbars}}^{-1}\bs{J}\right\}\bs{\nu}_{+1}^T\bs{\nu}_{+1} + 2(1-m_{\ts{e}}).
    \label{EQ17}
\end{align}
Next, by differentiating \eqref{EQ13} with respect to time, we obtain
\begin{align}
\dot{V}_{+1}(\bs{\qbar}_{\ts{e}},\bs{\nu}_{+1}) = 
&-\bs{\nu}^T_{+1}\bs{n}_\ts{e} -\bs{\nu}_{+1}^T\bs{K}_{\bs{\qbars}}^{-1}\bs{K}_{\bs{\omega}} \bs{\nu}_{+1} - 2\dot{m}_{\ts{e}}.
\label{EQ18}
\end{align}
Additionally, since $\sigma = +1$, it follows that  
\begin{align}
\dot{m}_{\ts{e}} = -\frac{1}{2}\bs{\nu}_{+1}^T\bs{n}_{\ts{e}} + \frac{1}{2} k_{\bs{n}} \bs{n}^T_\ts{e}\bs{n}_\ts{e},
\label{EQ19}
\end{align}
which we use to further simplify \eqref{EQ18}. Namely,

\begin{align}
\begin{split}
\dot{V}_{+1}(\bs{\qbar}_{\ts{e}},\bs{\nu}_{+1}) &= -\bs{\nu}_{+1}^T\bs{K}_{\bs{\qbars}}^{-1}\bs{K}_{\bs{\omega}} \bs{\nu}_{+1} - k_{\bs{n}} \bs{n}^T_\ts{e}\bs{n}_\ts{e} \\
&\leq -c_1\left(\bs{\nu}_{+1}^T\bs{K}_{\bs{\qbars}}^{-1}\bs{K}_{\bs{\omega}} \bs{\nu}_{+1} + k_{\bs{n}} \bs{n}^T_\ts{e}\bs{n}_\ts{e}\right) \\ 
& < 0,
\end{split}
\label{EQ20}
\end{align}
\mbox{$\forall\,\left\{ \bs{\qbar}_{\ts{e}},\bs{\nu}_{+1} \right\} \not\in \left\{ \bs{\qbar}_{\ts{e}}^{\star},\bs{\nu}_{+1}^{\star} \right\}
\cup
\{ \bs{\qbar}_{\ts{e}}^{\dagger},\bs{\nu}_{+1}^{\star} \}$}, where \mbox{$1 \leq c_1 \in \mathbb{R}$}. 

Thus, using \mbox{Theorem\,4.9} in~\cite{khalil2002nonlinear} and the conditions specified by~(\ref{EQ15})~and~(\ref{EQ20}), it can be determined that the equilibrium point \mbox{$\{\bs{\qbar}^{\star}_{\ts{e},+1}, \bs{\nu}^{\star}_{+1} \}$} is uniformly asymptotically stable in~\mbox{$\mathcal{D}_{+1} = \{\{ \bs{\qbar}_{\ts{e}},\bs{\nu}_{+1}\} \in \{\mathcal{S}^3,\mathbb{R}^3 \}~| V_{+1}(\bs{\qbar}_{\ts{e}}, \bs{\nu}_{+1})<4\}$}---a conservative estimate of the region of attraction---since using Lyapunov's indirect method, it can be shown that \mbox{$\{ \bs{\qbar}_{\ts{e}}^{\dagger},\bs{\nu}_{+1}^{\star}\}$} is an unstable saddle point and, additionally, noting that \mbox{$V_{+1}(\bs{\qbar}^{\dagger}_{\ts{e}}, \bs{\nu}^{\star}_{+1})=4$}. These findings indicate that all \textit{experimental} trajectories starting inside $\mathcal{D}_{+1}$ are attracted by \mbox{$\{ \bs{\qbar}_{\ts{e}}^{\star},\bs{\nu}_{+1}^{\star}\}$} and rejected by \mbox{$\{ \bs{\qbar}_{\ts{e}}^{\dagger},\bs{\nu}_{+1}^{\star}\}$}, a fact with important implications from a practical engineering perspective. Namely, because of the existence of \mbox{$\{ \bs{\qbar}_{\ts{e}}^{\dagger},\bs{\nu}_{+1}^{\star}\}$}, we cannot claim that \mbox{$\{ \bs{\qbar}_{\ts{e}}^{\star},\bs{\nu}_{+1}^{\star}\}$} is globally asymptotically stable; however, the CL system behaves as such because because whenever the system is initialized at \mbox{$\{ \bs{\qbar}_{\ts{e}}^{\dagger},\bs{\nu}_{+1}^{\star}\}$} or any of its converging trajectories, any \textit{\mbox{real-life}} disturbance, large or small, would move the system's state into a trajectory that approaches \mbox{$\{ \bs{\qbar}_{\ts{e}}^{\star},\bs{\nu}_{+1}^{\star}\}$} as \mbox{$t \rightarrow \infty$}.
\begin{figure*}[t!]
\vspace{1ex}
\begin{center}
\includegraphics[width=0.98\textwidth]{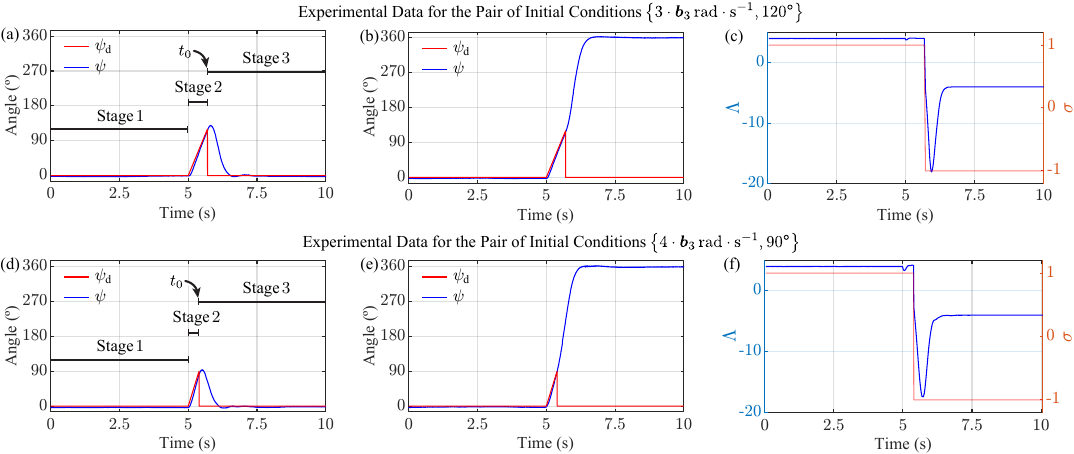}
\end{center}
\vspace{-2ex}
\caption{\textbf{Experimental data corresponding to two sets of flight tests}. \textbf{(a)}\,\mbox{Yaw-angle} reference, $\psi_{\ts{d}}$, and measured yaw angle, $\psi$, for the pair \mbox{$\left\{ \bs{\omega}_0, \psi_0 \right\} = 
\left\{3\cdot\bs{b}_3\,\ts{rad}\cdot\ts{s}^{-1},120^\circ \right\}$}, obtained with the benchmark controller. \textbf{(b)}\,\mbox{Yaw-angle} reference, $\psi_{\ts{d}}$, and measured yaw angle, $\psi$, for the pair \mbox{$\left\{ \bs{\omega}_0, \psi_0 \right\} = 
\left\{3\cdot\bs{b}_3\,\ts{rad}\cdot\ts{s}^{-1},120^\circ \right\}$}, obtained with the switching controller. \textbf{(c)}\,Instantaneous values of $\Lambda$ and $\sigma$ corresponding to the data shown in (b). \textbf{(d)}\,\mbox{Yaw-angle} reference, $\psi_{\ts{d}}$, and measured yaw angle, $\psi$, for the pair \mbox{$\left\{ \bs{\omega}_0, \psi_0 \right\} = 
\left\{4\cdot\bs{b}_3\,\ts{rad}\cdot\ts{s}^{-1},90^\circ \right\}$}, obtained with the benchmark controller. \textbf{(e)}\,\mbox{Yaw-angle} reference, $\psi_{\ts{d}}$, and measured yaw angle, $\psi$, for the pair \mbox{$\left\{ \bs{\omega}_0, \psi_0 \right\} = 
\left\{4\cdot\bs{b}_3\,\ts{rad}\cdot\ts{s}^{-1},90^\circ \right\}$}, obtained with the switching controller. \textbf{(f)}\,Instantaneous values of $\Lambda$ and $\sigma$ corresponding to the data shown in (e). \label{Fig03}}
\vspace{-3ex}
\end{figure*}

Now, for the subsystem corresponding to \mbox{$\sigma = -1$}, we show that the fixed point \mbox{$\{\bs{\qbar}^{\dagger}_{\ts{e},-1}, \bs{\nu}^{\star}_{-1}\}$} is asymptotically stable. In this case, we use the LF defined as
\begin{align}
V_{-1}(\bs{\qbar}_{\ts{e}}, \bs{\nu}_{-1}) = \frac{1}{2} \bs{\nu}_{-1}^T\bs{K}_{\bs{\qbars}}^{-1}\bs{J}\bs{\nu}_{-1} + 2(1 + m_\ts{e}),
\label{EQ21}
\end{align}
where, by definition, \mbox{$ \bs{\qbar}_{\ts{e}}(t) = \bs{\qbar}_{\ts{e},-1}(t)$} and, therefore, \mbox{$\bs{\qbar}^{\dagger}_{\ts{e}} = \bs{\qbar}^{\dagger}_{\ts{e},-1} = [-1\,\, 0\,\, 0\,\, 0]^T$}. Similarly to the case of $V_{+1}(\bs{\qbar}_\ts{e},\bs{\nu}_{+1})$, it can be shown that $V_{-1}(\bs{\qbar}_\ts{e},\bs{\nu}_{-1})$ is continuously differentiable, because both $\bs{\qbar}_\ts{d}$ and $\bs{\omega}_\ts{d}$ are smooth functions of time. Furthermore, by substituting \mbox{$\{\bs{\qbar}^{\dagger}_\ts{e},\bs{\nu}^{\star}_{-1}\}$} into (\ref{EQ21}), it follows that 
\begin{align}
V_{-1}(\bs{\qbar}_{\ts{e}}^{\dagger},\bs{\nu}_{-1
}^{\star})=0.
\label{EQ22}
\end{align}
Moreover, \mbox{$V_{-1}(\bs{\qbar}_\ts{e},\bs{\nu}_{-1})$} can be bounded according to
\begin{align}
    W_3(\bs{\qbar}_\ts{e},\bs{\nu}_{-1})\leq V_{-1}(\bs{\qbar}_\ts{e},\bs{\nu}_{-1}) \leq W_4(\bs{\qbar}_\ts{e},\bs{\nu}_{-1}),
    \label{EQ23}
\end{align}
with
\begin{align}
    W_3(\bs{\qbar}_\ts{e},\bs{\nu}_{-1})= \lambda_{\ts{min}}\left\{\bs{K}_{\bs{\qbars}}^{-1}\bs{J}\right\}\bs{\nu}_{-1}^T\bs{\nu}_{-1} + 2(1+m_{\ts{e}})
    \label{EQ24}
\end{align}
and
\begin{align}
    W_4(\bs{\qbar}_\ts{e},\bs{\nu}_{-1})= \lambda_{\ts{max}}\left\{\bs{K}_{\bs{\qbars}}^{-1}\bs{J}\right\}\bs{\nu}_{-1}^T\bs{\nu}_{-1} + 2(1+m_{\ts{e}}).
    \label{EQ25}
\end{align}
Next, by differentiating (\ref{EQ21}) with respect to time, we obtain
\begin{align}
\dot{V}_{-1}(\bs{\qbar}_{\ts{e}},\bs{\nu}_{-1}) =\, 
&\bs{\nu}^T_{-1}\bs{n}_\ts{e} -\bs{\nu}_{-1}^T\bs{K}_{\bs{\qbars}}^{-1}\bs{K}_{\bs{\omega}} \bs{\nu}_{-1} + 2\dot{m}_{\ts{e}}.
\label{EQ26}
\end{align}
Additionally, since \mbox{$\sigma = -1$}, it follows that  
\begin{align}
\dot{m}_{\ts{e}} = -\frac{1}{2}\bs{\nu}_{-1}^T\bs{n}_{\ts{e}} - \frac{1}{2} k_{\bs{n}} \bs{n}^T_\ts{e}\bs{n}_\ts{e},
\label{EQ27}
\end{align}
which we use to further simplify (\ref{EQ26}). Namely,
\begin{align}
\begin{split}
\dot{V}_{-1}(\bs{\qbar}_{\ts{e}},\bs{\nu}_{-1}) &= -\bs{\nu}_{-1}^T\bs{K}_{\bs{\qbars}}^{-1}\bs{K}_{\bs{\omega}} \bs{\nu}_{-1} - k_{\bs{n}} \bs{n}^T_\ts{e}\bs{n}_\ts{e}\\
&\leq -c_2\left(\bs{\nu}_{-1}^T\bs{K}_{\bs{\qbars}}^{-1}\bs{K}_{\bs{\omega}} \bs{\nu}_{-1} + k_{\bs{n}} \bs{n}^T_\ts{e}\bs{n}_\ts{e}\right)\\
& < 0,
\end{split}
\label{EQ28}
\end{align}
\mbox{$\forall\,\left\{ \bs{\qbar}_{\ts{e}},\bs{\nu}_{-1} \right\} \not\in \{ \bs{\qbar}_{\ts{e}}^{\dagger},\bs{\nu}_{-1}^{\star} \} \cup \left\{ \bs{\qbar}_{\ts{e}}^{\star},\bs{\nu}_{-1}^{\star}\right\}$}, where \mbox{$1 \leq c_2 \in \mathbb{R}$}. 

Thus, using \mbox{Theorem\,4.9} in~\cite{khalil2002nonlinear} and the conditions specified by \eqref{EQ23}~and~\eqref{EQ28}, it can be determined that the equilibrium point \mbox{$\{\bs{\qbar}^{\dagger}_{\ts{e},-1}, \bs{\nu}^{\star}_{-1} \}$} is uniformly asymptotically stable in 
\mbox{$\mathcal{D}_{-1} = \{\{ \bs{\qbar}_{\ts{e}},\bs{\nu}_{-1}\} \in \{\mathcal{S}^3,\mathbb{R}^3 \}~| V_{-1}(\bs{\qbar}_{\ts{e}}, \bs{\nu}_{-1})<4\}$}---a conservative estimate of the region of attraction---since using Lyapunov's indirect method, it can be shown that \mbox{$\{ \bs{\qbar}_{\ts{e}}^{\star},\bs{\nu}_{-1}^{\star}\}$} is an unstable saddle point and, additionally, noting that \mbox{$V_{-1}(\bs{\qbar}^{\star}_{\ts{e}}, \bs{\nu}^{\star}_{-1})=4$}. These findings indicate that all \textit{experimental} trajectories starting inside $\mathcal{D}_{-1}$ are attracted by \mbox{$\{\bs{\qbar}_{\ts{e}}^{\dagger},\bs{\nu}_{-1}^{\star}\}$} and rejected by \mbox{$\{\bs{\qbar}_{\ts{e}}^{\star},\bs{\nu}_{-1}^{\star}\}$}, a fact with important implications from a practical engineering perspective. Namely, because of the existence of \mbox{$\{ \bs{\qbar}_{\ts{e}}^{\star},\bs{\nu}_{-1}^{\star}\}$}, we cannot claim that \mbox{$\{ \bs{\qbar}_{\ts{e}}^{\dagger},\bs{\nu}_{-1}^{\star}\}$} is globally asymptotically stable; however, the CL system behaves as such because whenever the system is initialized at \mbox{$\{ \bs{\qbar}_{\ts{e}}^{\star},\bs{\nu}_{-1}^{\star}\}$} or any of its converging trajectories, any \textit{\mbox{real-life}} disturbance, large or small, would move the system's state into a trajectory that approaches \mbox{$\{ \bs{\qbar}_{\ts{e}}^{\dagger},\bs{\nu}_{-1}^{\star}\}$} as \mbox{$t \rightarrow \infty$}.

Having shown that the two subsystems specified by (\ref{EQ11}) have the same fixed points with reverse stability properties, now we show that the switching system is stable as a whole. First, we reiterate that for the subsystem corresponding to \mbox{$\sigma=+1$}, the stable pair is \mbox{$\bs{\qbar}_{\ts{e},+1}^{\star} = \left[ +1\,\, 0\,\, 0\,\, 0 \right]^T$} and \mbox{$\bs{\nu}_{+1}^{\star} =  \left[ 0\,\, 0\,\, 0 \right]^T\hspace{-0.4ex}$}, and the unstable pair is \mbox{$\bs{\qbar}_{\ts{e},+1}^{\dagger} = \left[ -1\,\, 0\,\, 0\,\, 0 \right]^T$} and \mbox{$\bs{\nu}_{+1}^{\star} =  \left[ 0\,\, 0\,\, 0 \right]^T\hspace{-0.4ex}$}; and, for the subsystem corresponding to \mbox{$\sigma=-1$}, the stable pair is \mbox{$\bs{\qbar}_{\ts{e},-1}^{\dagger} = \left[ -1\,\, 0\,\, 0\,\, 0 \right]^T$} and \mbox{$\bs{\nu}_{-1}^{\star} =  \left[ 0\,\, 0\,\, 0 \right]^T\hspace{-0.4ex}$}, and the unstable pair is \mbox{$\bs{\qbar}_{\ts{e},-1}^{\star} = \left[ +1\,\, 0\,\, 0\,\, 0 \right]^T$} and \mbox{$\bs{\nu}_{-1}^{\star} =  \left[ 0\,\, 0\,\, 0 \right]^T\hspace{-0.4ex}$}. Next, to apply \mbox{Theorem\,3.1} in~\cite{liberzon2003switching} to determine and enforce the stability of the CL switching system, we consider the difference between the LFs of the two subsystems specified by (\ref{EQ11}). Namely,
\begin{align}
\Delta V = V_{-1}  - V_{+1} = -2k_{\bs{n}}\bs{\omega}^T_\ts{e}\bs{K}_{\bs{\qbars}}^{-1}\bs{J}\bs{n}_\ts{e} + 4m_\ts{e}.
\label{EQ29}
\end{align}
By defining \mbox{$\Lambda = \Delta V$}, we guarantee that the value of the CL dynamics' LF decreases when the control law switches from \mbox{$\sigma=+1$} to \mbox{$-1$}. Similarly, when a switching from \mbox{$\sigma=-1$} to $+1$ occurs, the value of the CL dynamics' LF also decreases because in that case \mbox{$\Delta V \geq \delta$} and, therefore, \mbox{$V_{+1} - V_{-1} \leq -\delta$}. These facts imply that the change in value of the CL system's LF over time is always negative, even during switching events, because for both CL subsystems specified by (\ref{EQ11}), \mbox{$\dot{V}_{\sigma} < 0$}, for \mbox{$\sigma \in \left\{+1,-1\right\}$}, and the condition specified by (\ref{EQ29}) is always satisfied. \mbox{Consequently}, following the notation in~\cite{liberzon2003switching}, we can state that 
\begin{align}
V_{\sigma(t_l)} \left( \bs{\qbar}_{\ts{e}}(t_l), \bs{\nu}_{\sigma(t_l)}(t_l) \right) - V_{\sigma(t_j)} \left( \bs{\qbar}_{\ts{e}}(t_j), \bs{\nu}_{\sigma(t_j)}(t_j) \right) \leq -\delta,
\label{EQ30}
\end{align}
where $t_j < t_k < t_l$ and $\sigma(t_j) = \sigma(t_l) \neq \sigma(t_k)$.

In summary, since both CL subsystems specified by \eqref{EQ11} have asymptotically stable equilibrium points and, additionally, the condition in \eqref{EQ30} holds, asymptotic stability of the set \mbox{$\{\sigma \bs{\qbar}^{\star}_\ts{e}, \bs{\nu}^{\star}_{\sigma}\}$} follows from Theorem\,3.1 in~\cite{liberzon2003switching}.~~~~~~~~~~~~~~~~~~~~~~~~~~~~~~~~~~~~~~~~~~~~~~~~~~~~~~~~~~~~~~~~~~~~~~~~$\square$
\vspace{1.0ex}

\hspace{-2.2ex}\textbf{Remark\,1.} We selected as the switching function \mbox{$\Lambda = \Delta V$} considering both stability and performance. This design choice follows from noticing that the Lyapunov function of the CL system, $V_{\sigma}$, for \mbox{$\sigma \in \left\{+1,-1\right\}$}, is a measure of instantaneous energy and, therefore, the proposed switching controller was conceived to select the smallest instantaneous energy cost between $V_{+1}$ and $V_{-1}$. We defined the hysteretic rule specified by (\ref{EQ10}) to avoid chattering and excessive switching.
\vspace{1.0ex}
\begin{figure*}[t!]
\vspace{1ex}
\begin{center}
\includegraphics[width=0.98\textwidth]{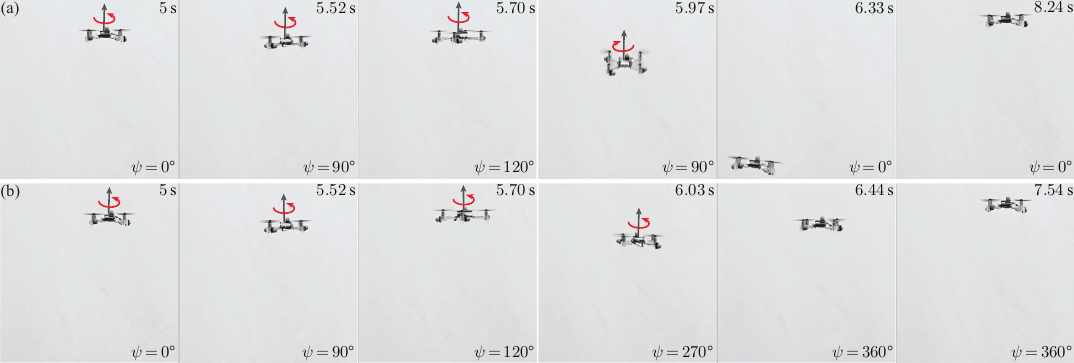}
\end{center}
\vspace{-2ex}
\caption{\textbf{Photographic sequences corresponding to two flight tests}. \textbf{(a)}\,Sequence corresponding to a flight test with the \mbox{initial-condition} pair \mbox{$\left\{ \bs{\omega}_0, \psi_0 \right\} = 
\left\{3\cdot\bs{b}_3\,\ts{rad}\cdot\ts{s}^{-1},120^\circ \right\}$} using the benchmark controller. \textbf{(b)}\,Sequence corresponding to a flight test with the \mbox{initial-condition} pair \mbox{$\left\{ \bs{\omega}_0, \psi_0 \right\} = 
\left\{3\cdot\bs{b}_3\,\ts{rad}\cdot\ts{s}^{-1},120^\circ \right\}$} using the switching controller. Video footage of these and other flight tests can be viewed in the accompanying supplementary movie. This movie is also available at \url{https://wsuamsl.com/resources/ROBOT2024movie.mp4}. \label{Fig04}}
\vspace{-3ex}
\end{figure*}

\section{Experimental Results}
\label{Section05}
\subsection{Performance Evaluation}
\label{Subsection05A}
To evaluate the flight performance obtained with the switching controller specified by (\ref{EQ09}) and compare it to that obtained using the benchmark controller specified by (\ref{EQ08}), we define a PFM that depends on the torque control signal, 
\begin{align}
\vspace{-1ex}
\Gamma_{\bs{\tau}} = \left(\frac{1}{t_{\ts{f}} - t_{0}}\int^{t_\ts{f}}_{t_{0}} \|\bs{\tau}_{\ts{exp}}(t)\|_2^2 dt\right)^{\frac{1}{2}},~~[\ts{N}\cdot \ts{m}]
\label{EQ31}
\end{align}
and a PFM that depends on the aerodynamic rotational power,
\begin{align}
\vspace{-1ex}
\Gamma_{\ts{p}} = \left(\frac{1}{t_{\ts{f}} - t_{0}}\int^{t_\ts{f}}_{t_{0}} \left[\bs{\tau}_{\ts{exp}}^T(t)\bs{\omega}(t) \right]^2 dt\right)^{\frac{1}{2}},~~[\ts{N}\cdot \ts{m}\cdot \ts{rad}\cdot\ts{s}^{-1}]
\label{EQ32}
\end{align}
where $t_{0}$ and $t_\ts{f}$ are the initial and final times of a flight maneuver, and \mbox{$\bs{\tau}_{\ts{exp}}\in\left\{\bs{\tau}_{\ts{b}}, \bs{\tau}_{\hspace{-0.2ex}\sigma}\right\}$} is the control law employed in a specific flight test. We selected $\Gamma_{\bs{\tau}}$ and $\Gamma_{\ts{p}}$ as PFMs instead of the CL system's LF because in this way we can compare different control methods using exactly the same metrics.

\subsection{Real-Time Flight Experiments}
\label{Subsection05C}
The \mbox{real-time} flight tests discussed in this section were performed using the experimental setup presented in~\cite{Goncalves2024ICRA}, which we briefly describe again here. The main component of the setup is an arena instrumented with a \mbox{six-V$5$-camera} Vicon \mbox{motion-capture} system running at $500\,\ts{Hz}$ and operated with the Tracker\,$3.9$ software package. During flight, we measure the six degrees of freedom and angular velocity of the controlled UAV using the Vicon system and an onboard \textit{inertial measurement unit} (\mbox{IMU,~BMI$088$}), respectively. After being processed by a PoE server of the Vicon system, the UAV's position and attitude are transmitted to it for control purposes using radio communication, enabled by the \mbox{Python-Crazyflie-client} library, and a Bitcraze antenna (Crazyradio\,PA). The proposed switching and benchmark attitude controllers are run onboard at a sampling rate of $500\,\ts{Hz}$; additionally, to avoid undesired translational drifting during flight, we control the UAV's position using the scheme presented in~\cite{Bena2023Yaw}, also running at a sampling rate of $500\,\ts{Hz}$. The instantaneous values of $\Lambda$ and $\sigma$, computed onboard by the switching controller, are transmitted to a host computer via radio for data collection at approximately $50\,\ts{Hz}$.

Experimental data corresponding to two sets of flight tests are presented in~Fig.\,\ref{Fig03}. These data are used to compare the results obtained with both the benchmark and proposed switching controllers. Each flight test discussed here is composed of the same three stages thoroughly described in~\cite{Goncalves2024ICRA} and indicated in Figs.\,\ref{Fig03}(a)~and~(d). As seen, the controlled UAV first hovers with a \mbox{yaw-angle} reference, $\psi_{\ts{d}}$, equal to zero (Stage\,$1$). Then, the UAV rotates rapidly along its yaw axis (Stage\,$2$), following a constant \mbox{angular-speed} reference, until the yaw reference is abruptly set to zero at time $t_0$ (the start of Stage\,$3$) with the initial condition \mbox{$\left\{\bs{\omega}(t_0),\psi\right(t_0)\} = \left\{\bs{\omega}_0,\psi_0\right\}$}, where $\psi$ is the measured yaw angle. Selecting \mbox{yaw-tracking} rotations instead of more complex flight maneuvers, such as multiple flips, allows us to readily evaluate functionality and performance. Note, however, that the tested control schemes can be used to execute any choice of rotational maneuver. We empirically selected the gains for the benchmark controller to be \mbox{$\bs{K}_{\bs{\qbars}} = 10^3\cdot\bs{J}\,\ts{N} \cdot \ts{m}$} and \mbox{$\bs{K}_{\bs{\omega}} = 10^2\cdot\bs{J}\,\ts{N} \cdot  \ts{m} \cdot \ts{s} \cdot \ts{rad}^{-1}$}, where \mbox{$\bs{J} = \ts{diag}\left\{16.6,16.7,29.3\right\}\cdot 10^{-6}\,\ts{kg}\cdot \ts{m}^2$}. For purposes of analysis and comparison, we selected the gains for the switching controller to be \mbox{$\bs{K}_{\bs{\qbars}} = 10\cdot\bs{J}\,\ts{N} \cdot \ts{m}$}, \mbox{$\bs{K}_{\bs{\omega}} = 10^2\cdot\bs{J}\,\ts{N} \cdot  \ts{m} \cdot \ts{s} \cdot \ts{rad}^{-1}$}, and \mbox{$k_{\bs{n}} = 10~\ts{rad} \cdot \ts{s}^{-1}$} because the resulting proportional and derivative weights for the two tested controllers, respectively defined by (\ref{EQ08})~and~(\ref{EQ09}), are almost identical---note that the term that depends on $\bs{\nu}_{\sigma}$ in  (\ref{EQ09}) contains an additional proportional weight.

\mbox{Figs.\,\ref{Fig03}(a)~and~(b)} show the data corresponding to the \mbox{yaw-angle} reference, $\psi_{\ts{d}}$, and measured yaw angle, $\psi$, for the pair \mbox{$\left\{ \bs{\omega}_0, \psi_0 \right\} = 
\left\{3\cdot\bs{b}_3\,\ts{rad}\cdot\ts{s}^{-1},120^\circ \right\}$}, obtained with the benchmark and switching controllers, respectively. \mbox{Fig.\,\ref{Fig03}(c)} shows the instantaneous values of $\Lambda$ and $\sigma$ corresponding to the data shown in \mbox{Fig.\,\ref{Fig03}(b)}. \mbox{Figs.\,\ref{Fig03}(d)~and~(e)} show the data corresponding to the \mbox{yaw-angle} reference, $\psi_{\ts{d}}$, and measured yaw angle, $\psi$, for the pair $\left\{ \bs{\omega}_0, \psi_0 \right\} = 
\left\{4\cdot\bs{b}_3\,\ts{rad}\cdot\ts{s}^{-1},90^\circ \right\}$, obtained with the benchmark and switching controllers, respectively. \mbox{Fig.\,\ref{Fig03}(f)} shows the instantaneous values of $\Lambda$ and $\sigma$ corresponding to the data shown in \mbox{Fig.\,\ref{Fig03}(e)}. In these two sets of experiments, the benchmark and switching controllers select different stable CL equilibrium AEQs after the \mbox{yaw-angle} reference is abruptly changed. Specifically, in these two particular cases, the benchmark scheme applies the torque computed according to $\bs{K}_{\bs{\qbars}}\bs{n}_{\ts{e}}$ in the direction of the shorter rotational path while the switching controller applies the torque computed according to $\bs{K}_{\bs{\qbars}}\bs{n}_{\ts{e},\sigma}$ in the direction of the longer rotational path. This dynamic behavior is explained by the fact that the benchmark controller only takes the instantaneous AEQ into account to compute the $\ts{sgn}\left\{m_\ts{e}\right\}$ while the switching controller uses both the instantaneous AEQ and \mbox{angular-velocity} error to evaluate the switching function specified by (\ref{EQ29}). \mbox{Fig.\,\ref{Fig04}} shows two photographic sequences corresponding to two flight tests performed using the \mbox{initial-condition} pair \mbox{$\left\{ \bs{\omega}_0, \psi_0 \right\} = 
\left\{3\cdot\bs{b}_3\,\ts{rad}\cdot\ts{s}^{-1},120^\circ \right\}$}. During the flight test shown in~Fig.\,\ref{Fig04}(a), the UAV was flown using the benchmark controller; during the flight test shown in~Fig.\,\ref{Fig04}(b), the UAV was flown using the proposed switching controller. The superior performance achieved by the switching controller becomes evident by noticing that the altitude of the UAV during flight was not significantly disturbed by abrupt changes in the \mbox{yaw-angle} reference.
 In contrast, the UAV drops noticeably when flown by the benchmark controller because it is forced to reverse its direction of rotation when the 
\mbox{yaw-angle} reference is abruptly varied. Video footage of these and other flight tests can be viewed in the accompanying supplementary movie. This movie is also available at \url{https://wsuamsl.com/resources/ROBOT2024movie.mp4}.

\begin{figure}[t!]
\vspace{1ex}
\begin{center}
\includegraphics[width=0.48\textwidth]{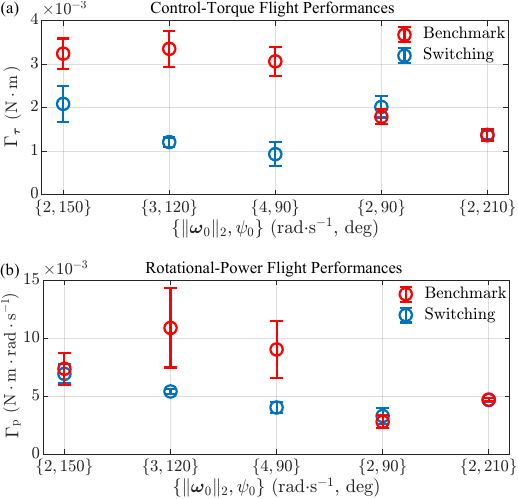}
\end{center}
\vspace{-2ex}
\caption{\textbf{Comparison of flight performances}. \textbf{(a)}~Means and ESDs of $\Gamma_{\bs{\tau}}$, for five different pairs of initial conditions. \textbf{(b)}~Means and ESDs of $\Gamma_{\ts{p}}$, for five different pairs of initial conditions. Each data point in these plots was computed from ten \mbox{back-to-back} experiments. \label{Fig05}}
\vspace{-3ex}
\end{figure}

To quantify and compare the performances obtained with the two tested controllers, we used the two PFMs specified by (\ref{EQ31}) and (\ref{EQ32}), respectively. We selected $t_0$ to coincide with the start of Stage\,$3$ and set \mbox{$t_{\ts{f}} = t_0+3$}. \mbox{Fig.\,\ref{Fig05}(a)} shows the computed means and \textit{empirical standard deviations} (ESD) of $\Gamma_{\bs{\tau}}$, for five different pairs of initial conditions. Each data point in this plot was computed from ten \mbox{back-to-back} experiments. \mbox{Fig.\,\ref{Fig05}(b)} shows the computed means and ESDs of $\Gamma_{\ts{p}}$, for five different pairs of initial conditions. Each data point in this plot was computed from ten \mbox{back-to-back} experiments. In these two plots, the first three \mbox{initial-condition} pairs correspond to cases in which the benchmark and switching controllers select different signs for the first term in their respective control laws---and, therefore, different stable CL equilibrium AEQs---and the last two pairs correspond to cases in which both controllers select the same signs for the first term in their respective control laws which, as expected, results in very similar PFM values. In the cases in which the two tested controllers select different stable CL equilibrium AEQs, the switching scheme performs notably better than the benchmark scheme. In fact, the worst experimental performance achieved by the switching controller is better than the best performance achieved by the benchmark controller in all tested cases regarding control effort and in two out of three cases regarding rotational power. Specifically, for the first three \mbox{initial-condition-pair} cases plotted in Fig.\,\ref{Fig05}, the proposed switching controller respectively reduces the \mbox{control-torque} and \mbox{rotational-power} PFM values by $49.75$\,\% and $28.14$\,\%, on average, with respect to those obtained with the benchmark controller. Last, it is worth mentioning that, in the flight tests in which different stable CL equilibrium AEQs are selected, the ESD values corresponding to the benchmark controller are, typically, significantly larger than those corresponding to the switching controller. This observation indicates that rotation reversions not only decrease the UAV's average flight performance but also the ability of the UAV to repeat some desired maneuvers.

\section{Conclusions and Future Work}
\label{Section06}
We presented a new method for synthesizing, analyzing, and implementing switching controllers capable of selecting the stable equilibrium AEQ of the CL attitude dynamics of a UAV during flight, according to an \mbox{energy-based} switching law. To analyze and enforce the stability of the CL switching dynamics, we used basic nonlinear Lyapunov theory. We tested and demonstrated the suitability, functionality, and performance of the proposed approach through \mbox{real-time} flight tests. The experimental results presented and discussed here are compelling as they clearly demonstrate that the proposed control approach can significantly improve the performance achieved by UAVs during \mbox{high-speed} rotational flight. The collected data show that in the experiments in which switching occurred, the values of two PFMs, respectively defined in terms of the control effort and rotational power, were reduced by $49.75\,\%$ and $28.14\,\%$, on average, when compared to those obtained with a benchmark controller. One limitation of the presented method is that we did not consider actuator saturation, which can greatly affect flight performance. This issue is a matter of current and future work in our laboratory.

\bibliographystyle{IEEEtran}
\bibliography{paper}


\end{document}